# Development of Yes/No Arabic Question Answering System

Wafa N Bdour[1], Natheer K. Gharaibeh[2]

[1]Alhoson College, Balqa Applied University, Irbed,00962, Jordan
`Wafa_bdour@bau.edu.jo`
[2]Ajloun College, Balqa Applied University, Irbed,00962, Jordan
`nkgharaibeh@bau.edu.jo`

*ABSTRACT*

*Developing Question Answering systems has been one of the important research issues because it requires insights from a variety of disciplines, including, Artificial Intelligence, Information Retrieval, Information Extraction, Natural Language Processing, and Psychology. In this paper we realize a formal model for a lightweight semantic–based open domain yes/no Arabic question answering system based on paragraph retrieval (with variable length). We propose a constrained semantic representation. Using an explicit unification framework based on semantic similarities and query expansion (synonyms and antonyms). This frequently improves the precision of the system. Employing the passage retrieval system achieves a better precision by retrieving more paragraphs that contain relevant answers to the question; It significantly reduces the amount of text to be processed by the system.*

## 1. INTRODUCTION

The growing interest in providing different information on the web has increased the need for a complicated search tool. Most existing information retrieval systems only provides documents, and this often makes users read a relatively large amount of full text [1].The study of question answering systems (QAS), which enable people to locate the information they need directly from large free-text databases by using their queries, has become one of the important aspects of information retrieval research[2].

Question Answering systems address the problem that while there is a huge amount of information in electronic format, there is no easy way of quickly and reliably accessing this information [3]. Using information retrieval systems (for example in Internet Search Engines) answer such questions indirectly to determine which documents probably contain the answer. It is then the role of the user to read those documents in order to find the answer. So Information Retrieval systems do not try to fully understand the meaning of users' questions and therefore often present as an "answer" a set of documents which are not relevant to the question, as usually happens with search engines [4].

Two basic types of question answering systems can be distinguished [5] : the first type : systems that attempt to answer a question by accessing structured information contained in a database, the main challenge in this type is to transform a natural language question into a database query. Often, systems of this type are also referred to as natural language interfaces to database systems, rather than stand-alone systems. Since database question answering systems use knowledge bases that are structured. The second type: systems that attempt to answer a question by analyzing unstructured information such as plain texts. Furthermore there are many actual systems that are hybrids of both types.

The pre-selection of documents that are considered for further analysis is a critical step in the whole question answering process. There is a tradeoff between selecting many or few documents; On the one hand, selecting too many documents might increase the computational costs to an extent which hurts the system's usefulness. It might also fail to reduce sufficiently noise, which may in turn hurt the performance of later modules in the question pipeline. [5] . On the other hand, selecting too few documents might have the effect that none of them contains an answer to the original question, while there are documents that do contain an answer.

Modern question Answering (QA) systems aim at providing answers to natural language questions in an open domain context. This task is usually achieved by combining information retrieval (IR) with information extraction (IE) techniques, modified to be applicable to unrestricted texts.[6]. For a question to be answered correctly, a QA system first has to understand what the question is asking about. Question answering (QA) systems have reached a remarkably high level of performance due to the integration of techniques from computational linguistic and information retrieval.

This is an important task of question task of question processing .Most current QA systems address it by identifying the type of answer sought. As General Question Answering (GQA) systems focus on WH-questions, many of which have named entities (NEs) as their answer, they usually classify answers according to different types of NE, such as product , organization , person [7] . WordNet is the main knowledge base that most current GQA systems use in analyzing relationships among words when calculating the similarity of a question and a candidate answer [7].

Much of the effort in QA until now has gone into building short answer QA systems, which answer questions for which the correct answer is a single word or short phrase (factoid questions) [8].Many questions are not in this class; they are better answered with a longer description or explanation. Producing these kinds of answers is the focus of long answer QA, an area still in early stages of development but already the subject of several recent pilot studies [8] .

The research issue at this paper is to identify the appropriate ways of ranking the documents in the collection with respect to yes/no question answering system. This allows the subsequent analysis steps to be restricted to a small number of documents, which allows for a more focused analysis. We selected this type of questions due to the accuracy and exact answers it give , furthermore it will be the best seed for ongoing research on the rest of question types.

The rest of this paper is structured as follows: Section two discusses related work. Section three describes a generic architecture for the Arabic Yes/No QA system and the new approach. Section four discusses testing and evaluation results of the new system. Section five contains our conclusions and future work to further improve our QA system.

## 2. LITERATURE REVIEW

The research in Question answering systems begin 1960s, Until 1990s, there were few research efforts in this area.[5]. Question answering systems that use natural language interfaces are not new [9], although older systems operated under limited domains .Wood's LUNAR (1977) was one of the first information retrieval systems to use NLP. The system attached a natural language

International Journal of Artificial Intelligence & Applications (IJAIA), Vol.4, No.1, January 2013International Journal of Artificial Intelligence & Applications (IJAIA), Vol.4, No.1, January 2013 interface to a database of geological samples. Users asked the system about information in the database, and the system responded by finding the answer in the database.

A number of systems have been designed to attempt question answering, usually involving specific domains. For example, BASEBALL QA system developed by Green et al. (1961) [10] which attempted to understand short narratives (in a restricted domain) and answer questions related to them.

A story understanding system (QUALM) is described by Lehnert (1981) It works through asking questions about simple, paragraph length stories .Part of the QUALM system includes a question analysis module that links each question with a question type. This question type guides all further processing and retrieval of information. [9]

Salton and McGill (1983) describe question answering (QA) systems as mainly provide direct answers to questions. Kupiec (1993) employed similar but rather simpler WH-question models to build a QA system. He used the interrogative words for informing the kinds of information required by the system [11].

In recent 90's , question answering achieved a great progress due to the Text Retrieval Conference (TREC)  , which has consisted of  a textual question answering session since 1999, with a wide range of research groups participating ,both from industry  and  academia [5] .

Diekema et el have developed a QA system .They classified a question type to different types: Wh-, yes/no, Alternative, Why, Definition, each has its own answering approach [12]. Question Answering in Wepcolpedia   have deal with yes/no question answering as a separate question type that investigate its own answering approach.

## 2.1 QAS in Arabic Language:

Research and development in the area of Arabic QA is not of high quality compared to similar work on English systems. In the context of the Arabic Question Answering (QA) task, Benajiba et al [13] produced ArabiQA which is a factoid centered Arabic QA system. The system was build using Java, it consists of a Passage Retrieval system, a Named Entities Recognition module and an Answer Extraction module, the result of the system report a precision of 83.3% over a manually created test dataset the details of which are not given.

Abouenour et el [14] presented a three-level approach for enhancing the Passage Retrieval (PR) stage ,the approach  use a semantic reasoning on top of keyword based and structure-based levels. Results of experiments conducted with a set of CLEF (Cross-Language Evaluation Forum) and TREC questions show an improvement of the Accuracy and the Mean Reciprocal Rank measures. An example shows also how the re-ranking based on the semantic score and Conceptual Graphs (CG) operations is even more relevant.

Kanaan et el [15] described Arabic QA system which makes use of  data redundancy rather than complicated linguistic analyses of either questions or candidate answers, to achieve its task. Akour et el [16] introduced a QA system (QArabPro) for Arabic Language. The system handles all types of questions including (How and Why). But generally similar to those used in a rule-based QA for English text , the authors uses a set of rules for each type of WH question.

In this research we want to focus on Yes or No Questions, to the best of our knowledge the research on Arabic Yes or No Questions fields have not been covered widely before.

## 3. SYSTEM ARCHITECTURE:

In the design of the QAS there are some factors [5] determines its components and modules : The context in which a question answering system is used, the anticipated user, the type of questions, the type of expected answers, and the format in which the available information is stored,

Our system is structured into three main modules: Question analysis module, Text retrieval module and Answer Selection module
1- Question Analysis module
2- Text retrieval module
3- Answer Selection module

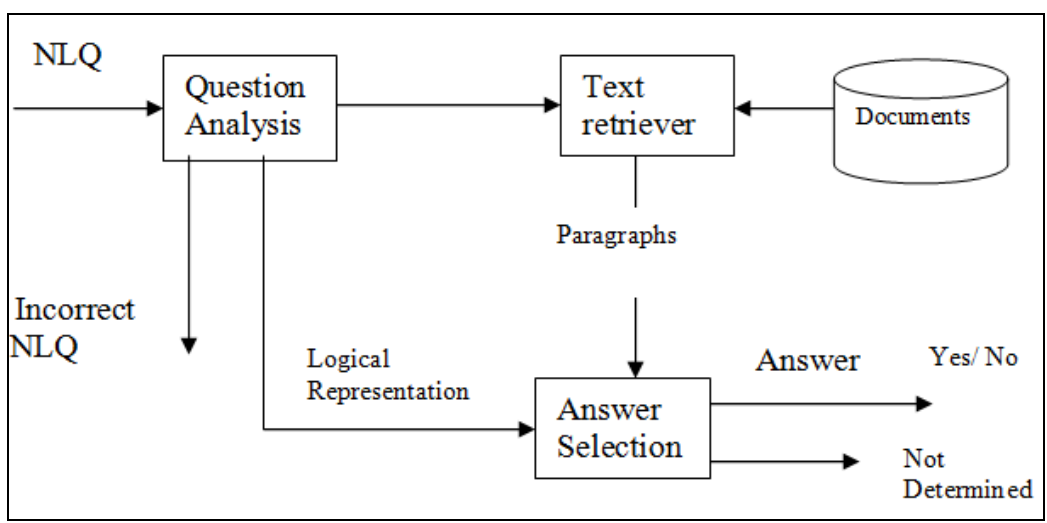

Figure 1: Schematic View of QAS

### 3.1 Question Analysis

When a Question is asked, the Question Analysis module's task does the following steps

1- Removing the question mark.
2- Removing the interrogative particle
3- Tokenizing: the tokenizer divides the user question into its separate words .And normalize the (Alef) letter.
4- Removing the stop words.
5- Removing the negation particles.
(if it exits) and set the negation property of the question representation
6- Tagging: we use tagger in order to determine the type of a word, verb or noun and obtain its root.
7- Parsing: recall that the Arabic sentence after the interrogative particle is nominal or verbal.

In nominal sentence, we are interested with the beginning noun "topic" (مبتدأ) which is the first noun after the interrogative particle (هل). And the comment noun (خبر) and we can mark it as the last noun without the article (ال).

In verbal sentence we are interested with the verb of the sentence which occur immediately after the interrogative particle (هل) , and the subject that follow the verb.

Using these rules and bottom up parsing algorithm, we can check if the question is well formed (correct), so generate the required logical representation with additional property if the question is negated or not, otherwise reject it [17].

To achieve the idea of query expansion, we retrieve a list of synonyms and antonyms for the verb in the verbal sentence. And to the comment in the nominal sentence and obtain their root lists. Using the thesaurus provided by the Microsoft word Arabic supported version.

In nominal sentence we are interested in the exact topic (مبتدأ) not its root so we only remove the article (ال) if exit – except (ال) in some words ,are stored in ALEF_LAM file) . In verbal sentence we are interested in the exact subject (فاعل) not its root so we also remove the article (ال) if exit ,

### 3.1.1 Logical Representation

At the core of our QAS is the Logical representation, which bridges the distinct representations of the functional structure obtained for questions and passages. we create 12 Logical representations for the Nominal and verbal Sentences as following:
With Nominal Sentences:

- Affirmative questions :
N (Topic, root (Comment), root ({remaining words }))
N (Topic, root (Comment Synonyms), root ({remaining words}))
~N (Topic, root (Comment Antonyms), root ({remaining words}))

- Negated questions :
~N (Topic, root (Comment), root ({remaining words}))
~N (topic, root (Comment Synonyms), root ({remaining words}))
 N (Topic, root (Comment Antonyms), root ({remaining words}))

With Verbal Sentences:

- Affirmative questions :
 V (Subject noun, root (verb), root ({remaining words}))
 V (Subject noun, root (verb Synonyms), root ({remaining words}))
 ~V (Subject noun, root (verb Antonyms), root ({remaining words}))

- Negated questions :
 ~V (Subject noun, root (verb), root ({remaining words}))
 ~ V (Subject noun, root (verb Synonyms), root ({remaining words}))
  V (Subject noun, root (verb Antonyms), root ({remaining words}))

Note that the negated of the negation question is affirmation.
Our suggested logical representation can be explained clearly by the following 4 examples.
- هل سميرة التي كسرت النافذة ؟
سميرة: مبتدأ
كسرت: خبر --> حطمت (Synonym)
So this example can be represented as :

N(سميرة, root (كسرت),root(النافذة))
N(سميرة, root (حطمت),root(النافذة))

Second example :

- هل محمد ولد جميل ؟

محمد: مبتدأ
جميل: خبر --> قبيح (Antonym)

This example can be represented as :
N(محمد, root (ولد),root(جميل))
~N(محمد, root (ولد),root(قبيح))

Third example :

- هل فتح محمود الباب ؟

فتح: فعل --> أغلق (Antonym)
محمود: فاعل

This example can be represented as :
V(محمود, root (فتح),root(الباب))
~V(محمود, root (أغلق),root(الباب))

fourth example :

- هل تكثر الأماكن السياحية في الأردن ؟

تكثر: فعل --> تزداد (synonym)
تكثر: فعل --> تقل (Antonym)
الأماكن: فاعل

This example can be represented as :

V(أماكن, root (تكثر), root(السياحية), root(الأردن))
V(أماكن, root (تزداد),root(السياحية),root(الأردن))
~V(أماكن, root (تقل),root(السياحية),root(الأردن))

There is another small subset of question structures which require preprocessing steps. Before we can represent it using the logical representation with verbal sentences. We can include them by checking the verb in the verbal sentences if its root match one of (وصف, شهر, ميز), this verb is replaced with the root of the world that attached at its beginning with the preposition (حرف الجر).

### 3.1.2 towards Discourse Knowledge:-

The representation just described based on the syntactic knowledge that can be extracted from Arabic sentences which determine the structural role of certain words. And to some extent the semantic knowledge of the sentence, using certain words meanings (their synonyms and antonyms if they exist) and how these meanings combine in sentence representations to form sentence meanings regardless of the context in which they used.

The representation of the con text-independent meaning of a sentence is called its logical form. Another advised knowledge level is the discourse knowledge which concerns how the immediately preceding sentence affects the interpretation of the next sentence. This information is especially important for interpreting pronouns. [18]

Arabic documents are full of implicit and explicit pronouns [19], so the best we can do is to look for the missing word in the preceding sentence, and hope that it was referred by the implicit or explicit pronoun in the first sentence. We named the implementation of this technique "Advanced search".

Examples:

- هل محمود الذي كسر النافذة ؟
  N ( محمود , root (حطم),root (النافذة))
  N ( محمود , root (كسر),root (النافذة))

According to the first representation , our advanced search can mark the following sentence as candidate answer.

- كان محمود يلعب كرة الطائرة , عندما حطم النافذة.
- قذف محمود الكرة باتجاه النافذة و فتحطمت .

## 3.2 Text Processing & Retrieval-:

They are 20 documents in our corpus. This module uses two techniques to retrieve the top 5 candidate paragraphs (with variable length (that are most relevant to the user question:

1- Paragraphs technique: - Split the documents into its built-in paragraphs and retrieve the top 5 paragraphs regardless from which document they are, according to some indexing scheme.

2- Document technique-:Retrieve the top 5 documents after they are ranked, then use the first indexing scheme to retrieve the top 5 paragraphs.

### 3.2.1 Paragraphs technique-:

- Each document is spitted its built-in paragraphs (with variable length(.

- Each paragraph is divided into its token (i.e. its words)
  Using the tokenize module and normalized the (ALEF) letter.

- Stop words are removed from each paragraph.

- For each token, we retrieve its root (term).

- Using the following formula , we can rank the paragraphs (p) according to its relevant to the question (q):
  Similarity $(p,q) = \sum W_{p,t} \cdot W_{q,t}$     (**Formula 1**)
  where     $W_{p,t} = (N/n) \log [(tf+1)/pl]$
  N is the total number of passages, n is the number of passages in which the term occurs, tf is the frequency of the term in the passage and pl, is the passage length (i.e. the number of non-stop terms in the passage).
  $W_{q,t} = (N/n) \log [(qtf+1)/ql]$
  qtf and ql are the frequency of the term within the query and the query length (not including stop terms) respectively.

- We order the paragraphs according to its similarity to the question in descending order and retrieve the top 5 paragraphs.

### 3.2.2 Document s Technique:-

1- Each paragraph is divided into its tokens (i.e. its words) using the tokenizer module and normalized the (ALEF) letter.

2- Stop words are removed from each paragraph.

3- For each token, we retrieve its root (term).

4- We rank the documents in the corpus using the following formula:
Similarity $(d_j,q) = \sum W_{i,j} \cdot W_i$  (Formula 2)
Where $W_{i,j} = F_{i,j} * idf_I = freq_{i,j} / max_k \{ freq_{k,j}\} * log_2 N/n_i$
$W_{i,q} = (.5 + (.5\ freq_{i,q}/ max \{freq_{k,j}\} ) * Log_2 N/n_i$
$freq_{i,j}$ : frequency of term $k_i$ in document $d_j$.
$freq_{i,q}$ : frequency of term $k_i$ in the query.
max $\{freq_{i,j}\}$ : the maximum frequency of any term in the documents.
N: the number of documents.
$n_i$ : the number of documents that the term $k_i$ appear in.

The term weight is given by fi,j * idfi

5-We put the documents in decreasing order according to the similarity value to the question.

6- Retrieve the top 5 documents.

7- Split those documents to their built-in paragraphs.

8-Order the paragraphs result from step 7 according to their similarity to the question in descending order using the formula used in the paragraphs technique and retrieve the top 5 paragraphs.

### 3.3 Answer Selection & generation:-

After the 5 paragraphs are selected using documents technique or paragraphs technique, we need to select the best sentence to represent the answer, and accordingly generates yes or no.
We follow these steps :

1- Split the paragraphs into their sentences .

2-In normal sentences we are interested in the exact topic (مبتدأ) not its used root, so we omit each sentence that does not contain it (in the original form ).
In verbal sentence we are interested in the exact subject (فاعل) not its used root , so we omit each sentence that does not contain it (in the original form ).

3-In the result sentence , we look for the remaining terms (in root form) that derived from the question in the logical representation (except the subject or the topic ), if the they exist , assign those indexes according to their position in the sentence. So each sentence will have its own rank as follow :
Rank =last occurrence - first occurrence

4- look for (أدوات النفي) negation particles in the selected answer (if exist).

International Journal of Artificial Intelligence & Applications (IJAIA), Vol.4, No.1, January 2013

5- Using the selected answer and the logical representation of the question to generate yes ,or no. a follows :

- Yes ,if : The question and the answer are affirmative .
  The question and the answer are negated.
- No, if :The question if affirmative and the answer are negated.
  The question is negated and the answer is affirmative.

## 4 EXPERIMENTS AND RESULTS

In order to test our project, we use a corpus of 20 Arabic documents, and a collection of 100 different yes/no question. We store these questions, with their correct answers generated manually . We answer these questions using our system. The results were as follows:

### 4.1 Using Documents Technique:

Using Documents technique without the suggested thesaurus , yields the following results.

Table 1:

| No. of documents Used | Correct Answers % | Incorrect Answers % | No of questions |
|---|---|---|---|
| 5 | 78% | 22% | 25 |
| 10 | 78% | 22% | 50 |
| 15 | 79% | 21% | 75 |
| 20 | 80% | 20% | 100 |

Using Documents technique with the thesaurus, yields the following results

Table 2:

| No. of documents used | Correct Answers % | Incorrect Answers % | No of questions |
|---|---|---|---|
| 5 | 79% | 21% | 25 |
| 10 | 80% | 20% | 50 |
| 15 | 81% | 19% | 75 |
| 20 | 83% | 17% | 100 |

Using Documents technique with the thesaurus, through the advanced search yields the following results.

Table 3:

| No. of documents used | Correct Answers % | Incorrect Answers % | No of questions |
|---|---|---|---|
| 5 | 81% | 19% | 25 |
| 10 | 82% | 18% | 50 |
| 15 | 84% | 16% | 75 |
| 20 | 85% | 15% | 100 |

International Journal of Artificial Intelligence & Applications (IJAIA), Vol.4, No.1, January 2013

The following figure abstract our results about documents technique, which shows the increasing in the correct answers percentage when the thesaurus and the advanced search are used especially with larger number of documents. The max percentage was 85% when 20 documents are used.

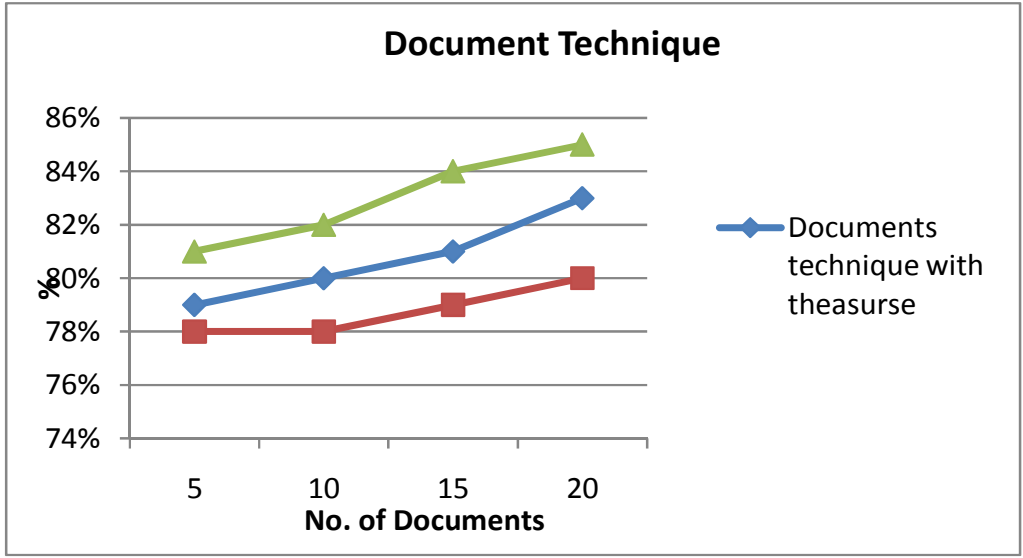

Figure 2 : The Documents Techniques

## 4.2 Using Paragraphs Technique:-

Using Paragraphs technique without the suggested thesaurus yields the following results.

Table 4:

| No. of documents Used | Correct Answers % | Incorrect Answers % | No of questions |
|---|---|---|---|
| 5 | 80% | 20% | 25 |
| 10 | 81% | 19% | 50 |
| 15 | 82% | 18% | 75 |
| 20 | 83% | 17% | 100 |

Using paragraphs technique with the thesaurus, yields the following results

Table 5:

| No. of documents used | Correct Answers % | Incorrect Answers % | No of questions |
|---|---|---|---|
| 5 | 78% | 22% | 25 |
| 10 | 78% | 22% | 50 |
| 15 | 79% | 21% | 75 |
| 20 | 80% | 20% | 100 |

Using paragraphs technique with the thesaurus, through the advanced search yields the following results.

International Journal of Artificial Intelligence & Applications (IJAIA), Vol.4, No.1, January 2013

Table 6 :

| No. of documents used | Correct Answers % | Incorrect Answers % | No of questions |
|---|---|---|---|
| 5 | 83% | 17% | 25 |
| 10 | 85% | 15% | 50 |
| 15 | 87% | 13% | 75 |
| 20 | 88% | 12% | 100 |

The following figure abstract our results about paragraphs technique, which shows the increasing in the correct answers percentage when the thesaurus and the advanced search are used especially with larger number of documents. The max percentage was 88% when 20 documents are used.

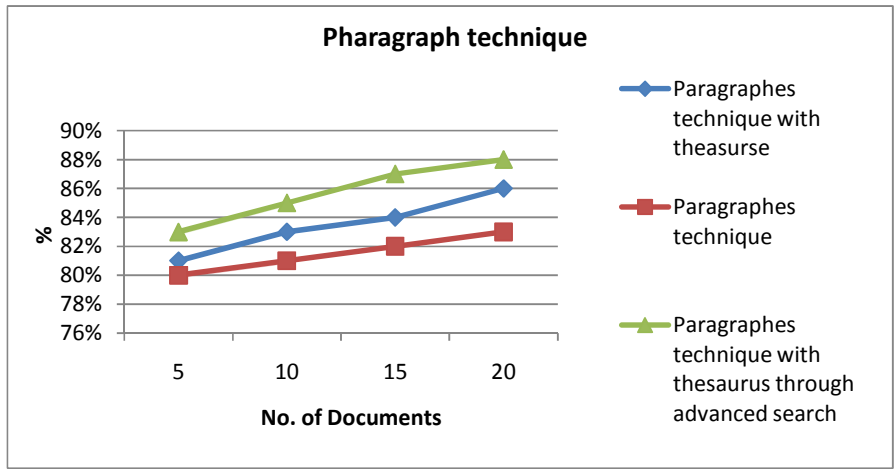

Figure 3 : The paragraphs Techniques

And the following figure shows the increasing percentage (3%) in the correct answers when paragraphs technique is used rather than documents technique, which occur especially when large number of documents are used.

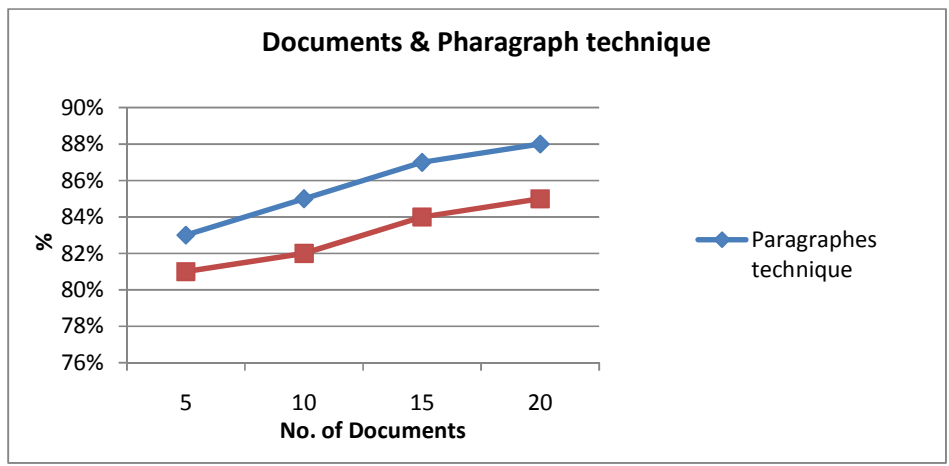

Figure 4 : The Paragraphes & Documents Techniques

## 6. CONCLUSIONS

Using a semantic logical representation based query expansion (synonyms and antonyms) frequently improve the precision of the system. Employing the paragraphs retrieval system achieves a better precision by retrieving more paragraphs that contain relevant answers to the question. And it significantly reduces the amount of text to be processed by the system and so the execution time.

A major percentage of 12% error in yes/no question answering system was due to the failure of the automatic tagger work that we used.

Such as two different words have the same root ,and the tagger give us two different roots for them. which let the matching process to be failed especially if one word was beginning with a preposition ب (حرف الجر ))? Also sometimes the correct result for determining the tag of the word : noun or verb which leads to significant mistakes.

Finally , incorrect results due to our work ,would occur because of the syntax of the user question, since we suppose that the question is the question is always abbreviated so that its extracted words (after removing the stop words) have to occur its candidate answers (except the syntactical structures we represented earlier.)

## 7. FUTURE WORK

Our research deals with yes/no questions, in which it answers with yes or no. Future work could be to provide extended responses that contain more that a plain "yes" or "no", by providing more specific or additional information (explanation) under which the answer is yes or no.

Another suitable future work in this area, could be to solve the problem of conditional responses. In which the question is answered with yes under certain conditions and no otherwise, that are missed from the question.

International Journal of Artificial Intelligence & Applications (IJAIA), Vol.4, No.1, January 2013